\documentclass[preprint,review,12pt]{elsarticle}

\usepackage[english]{babel}
\usepackage{blindtext}
\usepackage{amssymb}
\usepackage{graphicx}
\usepackage{lineno}
\usepackage{hyperref}
\usepackage{nomencl}
\usepackage[utf8]{inputenc}
\usepackage{bookmark}
\usepackage{lipsum}
\usepackage{amsthm}
\usepackage{epsfig}
\usepackage{amsmath}
\usepackage{cleveref}
\usepackage[final]{microtype}
\usepackage{textcomp}
\usepackage{caption}
\usepackage{subcaption}

\hypersetup{colorlinks = true,linkcolor = blue,anchorcolor =red,citecolor = blue,filecolor = red,urlcolor = red,
            pdfauthor=author}

\makenomenclature

\journal{arXiv}

\begin{document}

\begin{frontmatter}
	
	\title{Artificial Intelligence-Assisted Optimization and Multiphase Analysis of Polygon PEM Fuel Cells}
	
	\author[adrs]{Ali Jabbary \corref{cor1}}  
	\ead{st_a.jabbary@urmia.ac.ir}
	\author[adrs]{Nader Pourmahmoud}  
	
	\author[adrs]{Mir Ali Asghar Abdollahi}
    
    \author[adrs2]{Marc A. Rosen}
    
	\address[adrs]{Mechanical Engineering Department, Urmia University, Urmia, 5756151818, Iran}
	
	\address[adrs2]{Faculty of Engineering and Applied Science, University of Ontario Institute of Technology, Oshawa, Ontario, L1G 0C5, Canada}

	\cortext[cor1]{Corresponding Author}

	\begin{abstract}
	    This article presents new hexagonal and pentagonal PEM fuel cell models. The models have been optimized after achieving improved cell performance. The input parameters of the multi-objective optimization algorithm were pressure and temperature at the inlet, and consumption and output powers were the objective parameters. The output data of the numerical simulation has been trained using deep neural networks and then modeled with polynomial regression. The target functions have been extracted using the RSM (Response Surface Method), and the targets were optimized using the multi-objective genetic algorithm (NSGA-II). Compared to the base model, the optimized Pentagonal and Hexagonal models increase the output current density by 21.8\% and 39.9\%, respectively.
	\end{abstract}
	
	
	
	\begin{keyword}
		PEM fuel cell \sep Optimization \sep CFD \sep Neural network \sep Performance
	\end{keyword}
	
\end{frontmatter}


\section{Introduction}
\label{s:intro}
Conventional fuel sources cannot fulfill our energy demands today, and their usage pollutes the environment significantly \cite{akorede2010distributed}.
Renewable energy sources have gained reputations due to sustainability and fewer environmental side-effects \cite{owusu2016review}. Polymer Electrolyte Membrane Fuel Cells (PEMFCs) have been studied and used by many researchers due to their ability to generate electrical energy from chemical energy and oxidants \cite{wang2011review}.
PEM fuel cell is the most commonly utilized type of cell because of its quick starting time \cite{barbir2006pem} and convenience in transportation \cite{peighambardoust2010review}, and minor stationary energy requirements \cite{hamelin2001dynamic}. Furthermore, the ever-increasing use of PEM fuel cells needs more research to make it more promising \cite{banham2017current}.

The challenges that the PEM fuel cell faces are water and thermal management \cite{owejan2009water, berg2004water, ozturk2020investigation, rahimi2017design, liu2020anode, xu2020modelling, zhang2020design, yuan2020thermal}, performance optimization \cite{carcadea2020pem, lan2020analysis, cai2020design, wang2020ai}, design and modeling \cite{cai2020design, pan2020design, mojica2020experimental, abdollahi2022effect}, and cell humidification \cite{shao2020comparison, zhao2018study, wilberforce2019effect, liu2017modeling, subin2018experimental, cao2021pem}.
One of the critical topics that can be applied to achieve proper performance for the fuel cell is geometric design configuration \cite{wang2015theory}.

Bipolar plates (BP), gas channels, gas diffusion layers (GDL), catalyst layers (CL), and the polymer electrolyte membrane are the five essential components of a PEM fuel cell \cite{wang2011review}.
The membrane electrode assembly (MEA) is placed between the current collectors. MEA consists of five parts, including a polymer electrolyte membrane, CL, and GDL at each end \cite{carcadea2020pem}.

The cell's geometric design and flow field play an essential role in fundamental parameters such as how the species are diffused, velocity, temperature and pressure distributions, liquid water content, and current density production \cite{berning2002three}.
Considerable study has been conducted on the cell structure to enhance its performance \cite{manso2012influence}. Mass transfer, temperature diffusion, and electrochemical performance can be enhanced by initiating suitable flow fields and decreasing pressure drop in the cell \cite{li2005review, wilberforce2019comprehensive}.

Dong et al. investigated the energy performance of a PEM fuel cell by using discontinuous S-shaped and crescent ribs into flow channels \cite{dong2021improved}. Asadzade et al. simulated a new bipolar plate based on a lung-inspired flow field for PEM fuel cells to gain higher current densities \cite{asadzade2017design}. Seyhan et al. used artificial neural networks to forecast the cell performance with wavy serpentine channels \cite{seyhan2017performance}.

Afshari et al. analyzed a zigzag flow channel design for cooling PEM fuel cell plates, and their design provided better temperature control in the cell \cite{afshari2017numerical}. Jabbary et al. performed a three-dimensional numerical study on a PEM fuel cell with a rhombus design. The results of this study indicate that the use of this design significantly increases the power and current density \cite{jabbary2021numerical}. They conducted another study using a new cylindrical configuration on cell performance, and water flooding \cite{samanipour2020effects}. This design reduces the amount of liquid water in MEA levels and prevents water flooding.

A useful way to increase the performance and output power is geometric or parametric optimization \cite{wang2017parametric}. Optimizing the fuel cell's main parameters before the final operation is one of the most effective ways to reduce production and maintenance costs \cite{fletcher2016energy}. Researchers used numerous methods for mathematical optimization in the fuel cell field. Bio-inspired methods like particle swarm \cite{sarma2020design}, whale \cite{el2019semi}, genetic algorithm \cite{cai2020design}, Gray Wolf \cite{vatankhah2018optimally}, seagull \cite{lei2020power}, and fish swarm \cite{bai2017optimization} methods are among these.

Cao et al. experimentally analyzed PEM fuel cells using a new, improved seagull optimization algorithm \cite{cao2019experimental}. In addition, Miao et al. introduced a new optimization method called the Hybrid Gray Wolf Optimizer to obtain the optimal parameters of the PEM fuel cell \cite{miao2020parameter}.
Song et al. \cite{song2004numerical} examined one- and two-parameter numeric optimization analysis of the catalyst layer of the PEM fuel cell to maximize the current density of the catalyst layer with a given electrode potential.

The advantages of using the genetic algorithm are \cite{sivanandam2008genetic, dehghanian2009designing}:
\begin{itemize}
    \item Implementation of this theory is simple.
    \item It searches the population points, not a single point.
    \item It employs payout data rather than derivatives.
    \item It provides multi-target optimization.
    \item It does not utilize deterministic rules but employs probabilistic transitional rules.
\end{itemize}

Artificial intelligence (AI) can broadly be integrated as a fascinating modern technology with most research areas to solve challenges. As a result, this methodology has proved to have a high potential for advanced improvement in technological growth \cite{dwivedi2019artificial}. 
AI assists in the enhancement of performance and the development of new enterprise models \cite{wamba2020influence}. In addition, embedded AI solutions can be used to optimize manufacturing processes and extend machines and services with intelligent functions \cite{tao2018data}. As a result, artificial intelligence (AI) will be a critical component in the future competitiveness of mechanical engineering products and processes.

This study presents new pentagonal and hexagonal fuel cell designs. In multiphase analysis, optimization techniques based on artificial intelligence have been used to investigate the effects of critical parameters on current density and output/consumed powers. furthermore, multiple analyses given the average cell's pressure, temperature, velocity, and water content were performed to determine the performance of these models.
This study aims to achieve maximum output power while maintaining minimum consumed power.

\section{Methodology}
\label{s:method}
\subsection{Physical Model}
\begin{figure*}[!htb]
	\centering
	\subfloat[Cubic model]{\includegraphics[scale=0.17]{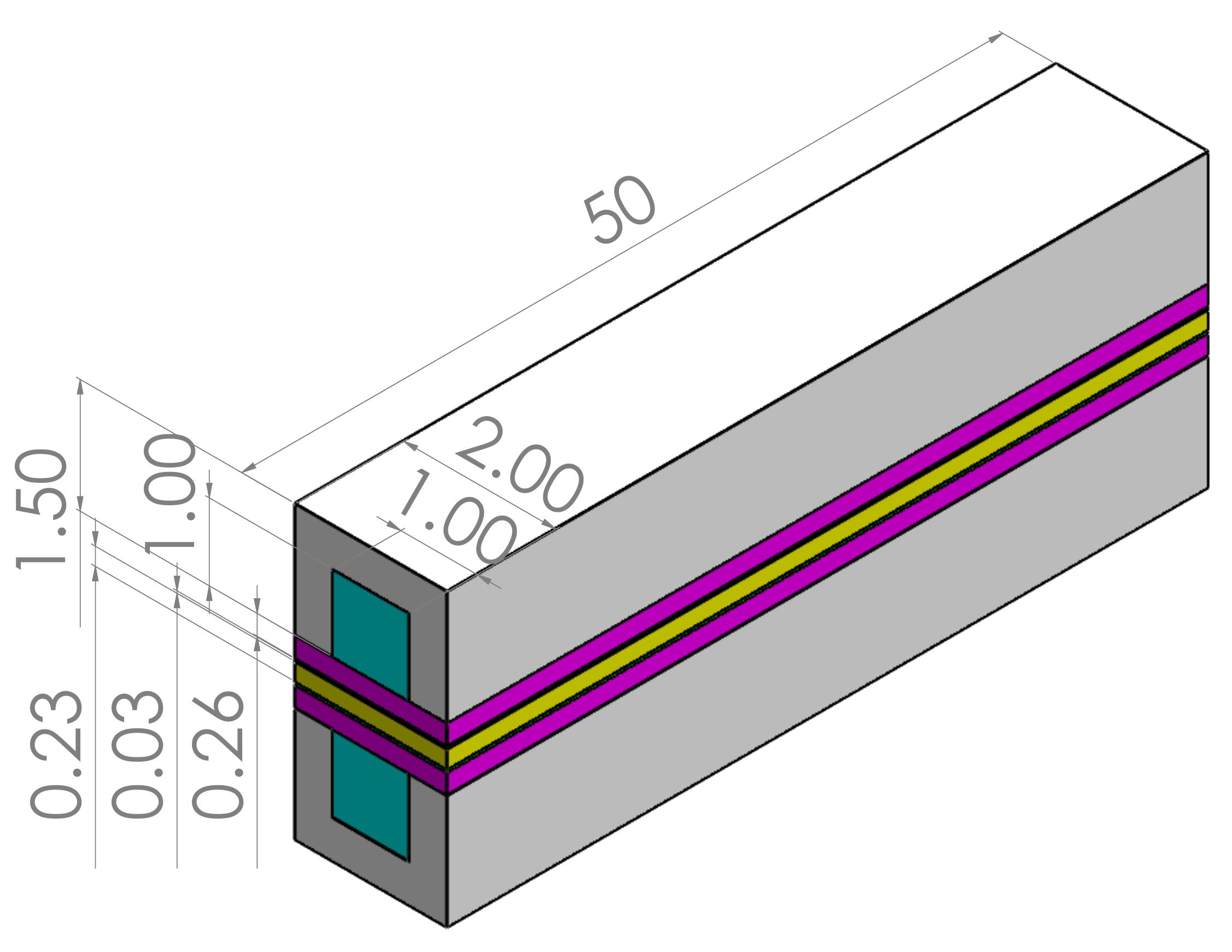}}
	\subfloat[Pentagonal model]{\includegraphics[scale=0.17]{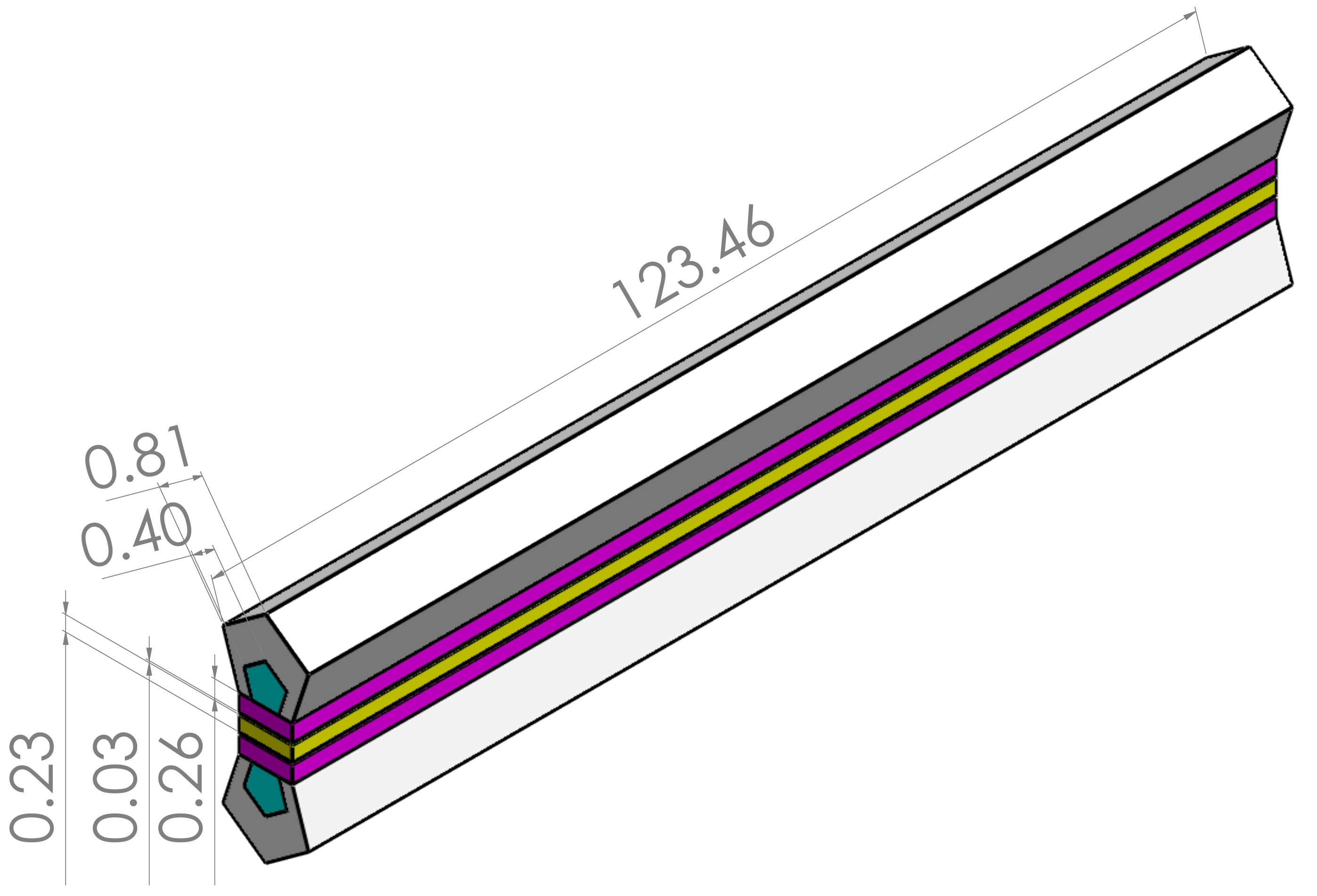}}
	\subfloat[Hexagonal model]{\includegraphics[scale=0.17]{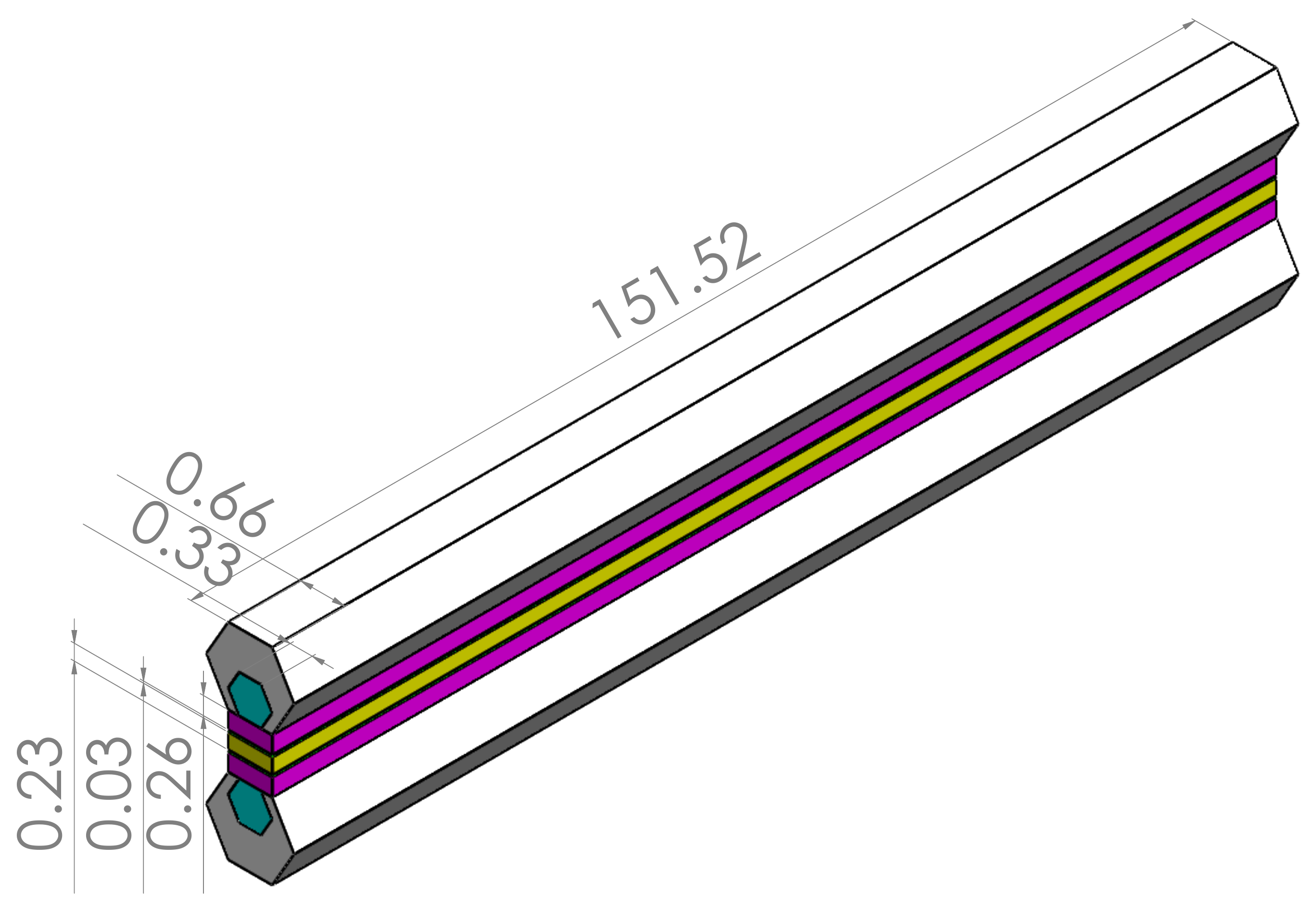}}
	\caption{PEM fuel cell models (sizes are in \(mm\))}
	\label{fig:models}
\end{figure*}

Physical and electrochemical phenomena were numerically investigated using the fuel cell model based on CFD techniques. The base design of the fuel cell (Cubic) and the polygon models used in this research are shown in Figure \ref{fig:models}, which shows the cell's main components.

The sizes and geometric specifications are displayed on this figure and are described in Table \ref{tab:geo}.

\begin{table*}[!htb]
\centering
\caption{Geometric specifications of the presented models}
\label{tab:geo}
\begin{tabular}{lllll}
\hline
\textbf{Parameter}          & \textbf{Unit}   & \textbf{Cubical} & \textbf{Pentagonal} & \textbf{Hexagonal} \\ \hline
Channel side       & mm     & 1       & 0.4        & 0.3      \\
Channel length     & mm     & 50      & 123.5     & 151.5    \\
BP side            & mm     & 1.5|2 & 0.81       & 0.66      \\
GDL thickness      & mm     & 0.26    & 0.26       & 0.26      \\
CL thickness       & mm     & 0.03    & 0.03       & 0.03      \\
Membrane thickness & mm     & 0.23    & 0.23       & 0.23      \\
MEA wet area       & $\text{mm}^2$ & 100     & 100    & 100   \\
Inlet/Outlet area  & $\text{mm}^2$ & 1       & 0.28       & 0.28      \\ \hline
\end{tabular}
\end{table*}

\subsection{Model Assumptions}
A \textit{comprehensive fuel cell} is a highly complex device that includes fluid dynamics, mass transport phenomena, and electrochemical processes. To analyze a problem involving a three-dimensional model, the following simplification assumptions must be made \cite{wang2004fundamental}:
\begin{itemize}
    \item Channel flows are assumed to be laminar, incompressible, and steady;
    \item PEMFC operates in non-isothermal, multiphase, and steady-state situations;
    \item GDL and CL are isotropic and homogenous;
    \item The MEA (membrane electrode assembly) is homogenous porous media with uniform porosity;
    \item The membrane is entirely humidified, ensuring consistent ionic conductivity.
    
\end{itemize}

\subsection{Governing Equations}
Reactive flow, species transfer, reactive consumption, water production, temperature and pressure diffusion are only few of the factors that must be studied in order to evaluate cell performance. Different variables are studied using a three-dimensional steady-state model.

\subsubsection{Continuity equation}
The conservation of mass equation (continuity) in all regions is \cite{atyabi2019three}:
\begin{equation}
    \frac{\partial u}{\partial x}
    + \frac{\partial v}{\partial y}
    + \frac{\partial w}{\partial z}
    = 0
\end{equation}

Electrodes are manufactured by carbon cloth, or carbon fiber \cite{zhou2016highly}. Therefore, they are recognized as a porous medium where distributed reactant gases (species). Concerning the porosity of MEA ($\epsilon$), the continuity equation is given by:

\begin{equation}
    \frac{\partial(\rho \epsilon u)}{\partial x}
    + \frac{\partial(\rho \epsilon v)}{\partial y}
    + \frac{\partial(\rho \epsilon w)}{\partial z}
    = S_m
\end{equation}

\noindent$u$, $v$, and $w$ are velocity in the $x$, $y$, and $z$ axes, individually, $\rho$ is the density of reactant gases. $S_m$ is the mass sink/source term, and it is considered zero since no reaction occurs in the flow channels and GDLs. However, because of the reactivity of reactant species, the sink/source term is not zero in the catalyst layer and can be evaluated by \cite{toghyani2018thermal}:

\begin{equation}
    S_{\text{H}_2} = -\frac{M_{\text{H}_2}}{2F} R_{a}
\end{equation}

\begin{equation}
    S_{\text{O}_2} = -\frac{M_{\text{O}_2}}{4F} R_{c}
\end{equation}

\begin{equation}
    S_{\text{H}_\text{2}\text{O}} = +\frac{M_{\text{H}_\text{2}\text{O}}}{2F} R_{c}
\end{equation}

\noindent $F$ is the Faraday constant ($96,485 \frac{C}{mol}$), and $M$ is the species' molecular weight ($\frac{kg}{mol}$) and $R$ can be calculated by Butler–Volmer equations (\ref{eq:bv1} and \ref{eq:bv2}).

\subsubsection{Momentum Equations}
The momentum conservation equations:

\begin{equation}
    \nabla . (\rho \vec{u} \vec{u})
    = -\nabla P + \nabla . (\mu \nabla \vec{u}) + S_{p, i}
\end{equation}

\noindent $S_{p, i}$ is the sink/source term for porous media in the $x$, $y$, and $z$ axes. As the pressure decreases in porous media, Darcy's law has been estimated in the model. The source term is \cite{das2018equations}:

\begin{equation}
    S_{p, i} = -\left(\sum_{i=1}^3 \frac{1}{\beta_i} \mu u_i \right)
\end{equation}

\noindent $\beta$ is the permeability of the media.

\subsubsection{Energy Conservation}
The energy conservation equation:

\begin{equation}
    \nabla . \left(\rho C_p \vec{u} T \right) =
    \nabla . \left(k_{\text{eff}} \nabla T \right) + S_h
\end{equation}


\noindent where \(k_{\text{eff}}\) denotes effective thermal conductivity, \(C_p\) denotes specific heat at constant pressure, and \(S_h\) is the additional volumetric source term in the energy equation. \(k_{\text{eff}}\) is calculated by:

\begin{equation}
    k_{\text{eff}} = \epsilon k_f + (1 - \epsilon) k_s
\end{equation}

\noindent And \(S_h\) is evaluated by:

\begin{equation}
    S_h = h_{\text{react}} - R_{an, cat} \eta_{an, cat} + I^2 R_{ohm} + h_L
\end{equation}

\noindent $h_{\text{react}}$ represents the net change in enthalpy caused by electrochemical reactions. The ohmic term is estimated because there is no heat production operator in bipolar plates. As a result, the energy equation is reduced to:

\begin{equation}
    \nabla . (k \nabla T) = -R_{ohm} I^2
\end{equation}

\noindent $k$ is the conductivity of the bipolar plates.

\subsubsection{Mass Transfer (Species Transport Equations)}
The reactant gases are hydrogen and air, which can be assumed to behave as ideal gases. The following are the equations for species transport:

\begin{equation}
    \nabla . \left(\epsilon \vec u C_i \right) =
    \nabla . \left(D_i^{eff} \nabla C_i \right)
    + S_i
\end{equation}

\noindent $C_i$ is the species' molar concentration. $S_i$ is the extra volumetric source term of species such as $\text{H}_2$, $\text{O}_2$, and $\text{H}_2\text{O}$ for CLs zones and are calculated by;

\begin{equation}
    S_{\text{H}_2} = -\frac{R_a}{2F}
\end{equation}

\begin{equation}
    S_{\text{O}_2} = -\frac{R_c}{4F}
\end{equation}

\begin{equation}
    S_{\text{H}_\text{2}\text{O}} = \frac{R_c}{2F}
\end{equation}

In addition, the gas diffusivity coefficient ($D_i^{\text{eff}}$), which is determined by operation conditions, is provided by;

\begin{equation}
    D_i^{\text{eff}} = \epsilon^{1.5} (1-s)^{2.5}
    D_i^{\text{0}} \left(\frac{P_0}{P} \right)
    \left(\frac{T}{T_0} \right)^{1.5}
\end{equation}

\noindent $r_s$ is the saturation exponent of pore blockage, $D_i^{\text{0}}$ is the reference mass diffusivity of the $i_{\text{th}}$ species under standard conditions, and $s$ is water saturation (the volume percentage of liquid water) and is derived as follows;

\begin{equation}
    s = \frac{V_{\text{liquid}}}{V_{\text{total}}}
\end{equation}

\noindent $V$ is the volume.

\subsubsection{Butler–Volmer Equation}
The Butler-Volmer equation can define the volumetric transfer currents are as follows;

\begin{equation}
    \label{eq:bv1}
    R_a = \left(\zeta_a j_a^{\text{ref}} \right)_a
    \left(\frac{C_{\text{H}_2}}{C_{\text{H}_2}^{\text{ref}}} \right)^{\gamma_a}
    \left(e^{\frac{\alpha_a F \eta_a}{RT}}
    - e^{\frac{-\alpha_c F \eta_a}{RT}} \right)
\end{equation}

\begin{equation}
    \label{eq:bv2}
    R_c = \left(\zeta_c j_c^{\text{ref}} \right)_c
    \left(\frac{C_{\text{O}_2}}{C_{\text{O}_2}^{\text{ref}}} \right)^{\gamma_c}
    \left(-e^{\frac{\alpha_a F \eta_c}{RT}}
    + e^{\frac{-\alpha_c F \eta_c}{RT}} \right)
\end{equation}

\noindent The values $j^{\textit{ref}}$, $\zeta$, and $\alpha$ represent the reference exchange current density, specific active surface area, and transfer coefficient. In addition, $c$, $c_{\textit{ref}}$, and $\gamma$ represent the concentration of reactant flow, the reference value, and the concentration dependency, respectively.

\subsubsection{Charge Conservation Equations}
Electrochemical processes take place at the catalyst layers in PEM fuel cells. Surface activation overpotential is the main factor behind these responses \cite{sivertsen2005cfd}. Therefore, the potential difference between the solid and the membrane is referred to as the activation overpotential \cite{das2008three}. As a consequence, two charge equations are required. One equation for electron transport via conductive solid phase and another for proton transport across the membrane:

\begin{equation}
    \nabla . \left(\sigma_{sol} \nabla \phi_{sol} \right)
    + R_{sol} = 0
\end{equation}

\begin{equation}
    \nabla . \left(\sigma_{mem} \nabla \phi_{mem} \right)
    + R_{mem} = 0
\end{equation}

Current density \((A/m^3)\) is used to describe volume sink terminology. Only in the catalytic layers are these expressions set. For the solid phase, they are calculated by:

Anode side:
\begin{equation}
    R_{sol} = -R_a \quad (<0)
\end{equation}

Cathode side:
\begin{equation}
    R_{sol} = +R_c \quad (>0)
\end{equation}

For the membrane phase, they can be evaluated by the following equations:

Anode side:
\begin{equation}
    R_{mem} = +R_a \quad (>0)
\end{equation}

Cathode side:
\begin{equation}
    R_{mem} = -R_c \quad (<0)
\end{equation}

The following equation is used to determine average current density:

\begin{equation}
    i_{ave} = \frac{1}{A} \int_{V_a} {R_a} dV
    = \frac{1}{A} \int_{V_c} {R_c dV}
\end{equation}

\subsubsection{Water Transport via Membrane}
Water produced by the cathodic process in PEM fuel cells diffuses to the anode side. It will transport across the membrane via electro-osmosis force and back diffusion \cite{das2010analysis}. $\lambda$ is determined as the number of water molecules divided by the number of charged \(\text{HSO}_3\) sites. Springer et al. \cite{springer1991polymer} developed a formula for estimating it:

\begin{equation}
    \lambda = 
    \begin{cases}
    	0.043 + 17.81a - 39.85 a^2 + 36 a^3 & a < 1 \\
    	14 + 1.4 (a - 1) & a > 1
    \end{cases}
\end{equation}

$a$ denotes the water activity.

\subsubsection{Consumption and Production Powers}
The consumption and production powers are two important parameters affecting the performance of the PEM fuel cell. In section \ref{s:result}, the specified values are obtained and examined. The following equations can be used to determine these parameters: \cite{jabbary2021numerical}:
\begin{itemize}
    \item Production power:
    \begin{equation}
        P_{pro} = I.V.A_{eff}
    \end{equation}
    
    \item Consumption power:
    \begin{equation}
        P_{cons} = \Delta P.A_{in}.u_{in}
    \end{equation}
\end{itemize}
Here, \(A_{eff}\) is the effective area of the membrane and, \(A_{in}\) is the inlet area of the channel.
\subsection{Simulation Conditions}
Table \ref{tab:opt} demonstrates the initial operating conditions utilized for numerical simulation providing the same initial operating conditions to all models.

\begin{table}[!htb]
\centering
\caption{Operating conditions of the fuel cell models}
\label{tab:opt}
\begin{tabular}{lll}
\hline
\textbf{Parameter}        & \textbf{Unit} & \textbf{Value} \\ \hline
Operating pressure        & Pa           & 101325         \\
Operating temperature     & K             & 353.15         \\
Anode relative humidity   & $\%$            & 100        \\
Cathode relative humidity & $\%$            & 100        \\
Anode stoichiometry       & --            & 1.2            \\
Cathode stoichiometry     & --            & 2              \\ \hline
\end{tabular}
\end{table}

The boundary conditions of the models are given in Table \ref{tab:boundary}, and the specifications of the MEA layers are shown in Table \ref{tab:mea} according to Hashemi's study \cite{hashemi2012cfd}.

\begin{table*}[!htb]
	\centering
	\caption{Boundary conditions of the fuel cell models}
	\label{tab:boundary}
	\resizebox{\textwidth}{!}{
	\begin{tabular}{lllll} 
		\hline
		\textbf{Parameter}                                         & \textbf{Unit} & \textbf{Cubical} & \textbf{Pentagonal} & \textbf{Hexagonal} \\ 
		\hline
		Anode mass flow rate                                       & $\text{kg/s}$        & 1.3e-07        & 1.3e-07           & 1.3e-07          \\
		Cathode mass flow rate                                     & $\text{kg/s}$         & 1.4e-06        & 1.4e-06           & 1.4e-06          \\
		$\text{H}_2$ mass fraction at anode inlet                  & –           & 0.113            & 0.113               & 0.113              \\
		$\text{H}_\text{2}\text{O}$ mass fraction at anode inlet   & –           & 0.886            & 0.886               & 0.886              \\
		$\text{O}_2$ mass fraction at cathode inlet                & –           & 0.151            & 0.150               & 0.151              \\
		$\text{H}_\text{2}\text{O}$ mass fraction at cathode inlet & –           & 0.353            & 0.353               & 0.353              \\
		Inlet pressure                                             & atm           & 1                & 1                   & 1                  \\
		Relative inlet humidity                                    & $\%$           & 100          & 100             & 100            \\
		Inlet temperature at anode/cathode                         & K             & 353.15           & 353.15              & 353.15             \\
		\hline
	\end{tabular}}
\end{table*}

\begin{table*}[!htb]
	\centering
	\caption{Membrane Electrode Assembly (MEA) properties}
	\label{tab:mea}
	\resizebox{\textwidth}{!}{
	\begin{tabular}{llll}
		\hline
		\textbf{Parameter}                           & \textbf{Symbol}                 & \textbf{Unit} & \textbf{Value} \\ \hline
		GDL porosity                                 & \(\epsilon_{GDL}\)              & --            & 0.5            \\
		CL porosity                                  & \(\epsilon_{CL}\)               & --            & 0.5            \\
		Membrane porosity                            & \(\epsilon_{mem}\)              & --            & 0.6            \\
		Electrical conductivity of electrode         & \(\sigma_{sol}\)                & \(\text{S/m}\)       & 100            \\
		Proton conductivity of membrane              & \(\sigma_{mem}\)                & \(\text{S/m}\)       & 17.1        \\
		Thermal conductivity of electrode            & \(k_{\text{eff}}\)              & \(\text{W/mK}\)     & 1.3            \\
		Anode apparent charge transfer coefficient   & \(\alpha_{\text{an}}\)          & --            & 0.5            \\
		Cathode apparent charge transfer coefficient & \(\alpha_{\text{cat}}\)         & --            & 1.0            \\
		Anode exchange current density               & \(R_{\text{an}}^{\text{ref}}\)  & \(\text{A}/\text{m}^2\)    & 30           \\
		Cathode exchange current density             & \(R_{\text{cat}}^{\text{ref}}\) & \(\text{A}/\text{m}^2\)     & 0.004          \\ \hline
	\end{tabular}}
\end{table*}

\subsection{Numerical Procedure}
\begin{figure}[!htb]
    \centering
    \includegraphics[scale=0.7]{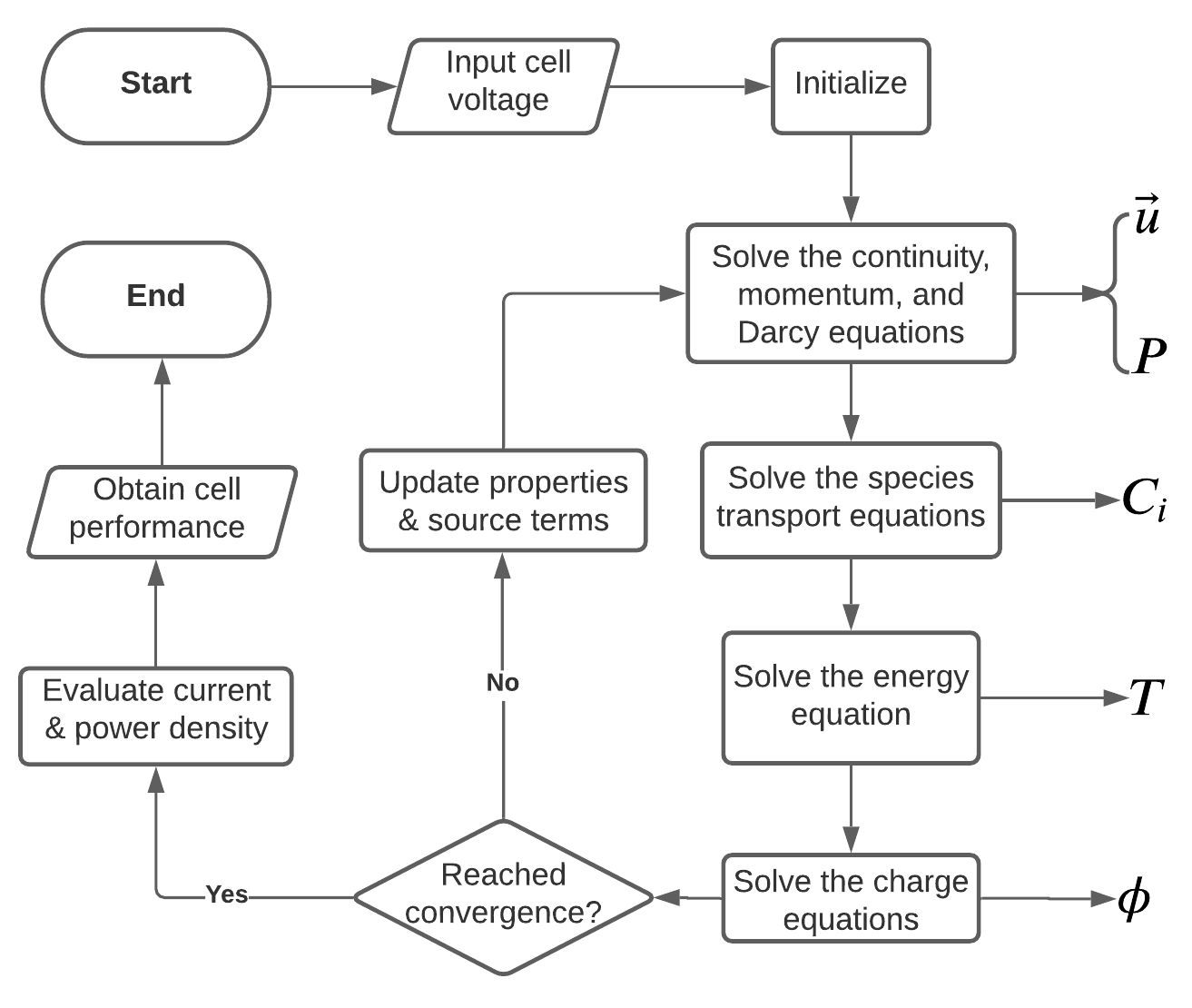}
    \caption{CFD algorithm of PEM fuel cell simulation}
    \label{fig:cfd}
\end{figure}

Figure \ref{fig:cfd} displays the presented CFD algorithm of PEM fuel cell simulation. ANSYS\textsuperscript{\textregistered} Fluent 2021 software was used in CFD analysis to solve the governing equations through computational domain using finite volume method.

The double-precision technique and the second-order upwind method were applied to discretize the terms. In addition, the multigrid \textit{F-Cycle} type \cite{mulder1989new} and the \textit{BCGSTAB} method (Biconjugate Gradient Stabilized Method) with 50 max course cycles were employed to stabilize the solutions and avoid divergence due to the complex nature of the governing equations. 

\textit{Stopping criteria} are requirements that must be met for the algorithm to be stopped. Considering that an iterative approach computes successive approximations to a nonlinear system's solution, a test is required to decide when to terminate the iteration. Stopping criteria would evaluate the distance between the latest iteration and the correct answer. These distances are called residuals. The lower the value of residuals, the closer the numerical analysis results to the existing solutions with fewer errors. When the residuals reach the desired value, the iteration is over, and the final results will be obtained.

\subsection{Artificial Intelligence-Assisted Optimization}
Mathematical optimization is an effective method for resolving complex problems by utilizing the most efficient resources and data. Optimization is the process of determining the values of decision variables to achieve a problem's goal. One of the most important applications of artificial intelligence is lowering the computing costs of optimization. An optimization model comprises appropriate objectives, variables, and constraints. The most reliable solution is choice variables that maximize or minimize the objective function while remaining within the solution range. The objectives of this study are the produced and consumed powers. \textit{Feed-forward deep neural networks} were used to model the data. \textit{Response Surface Method (RSM)} was applied for function approximation to extract objective functions and use them in the optimization algorithm.
\subsubsection{Machine Learning Model}
\begin{figure}[!htb]
    \centering
    \includegraphics[scale=0.8]{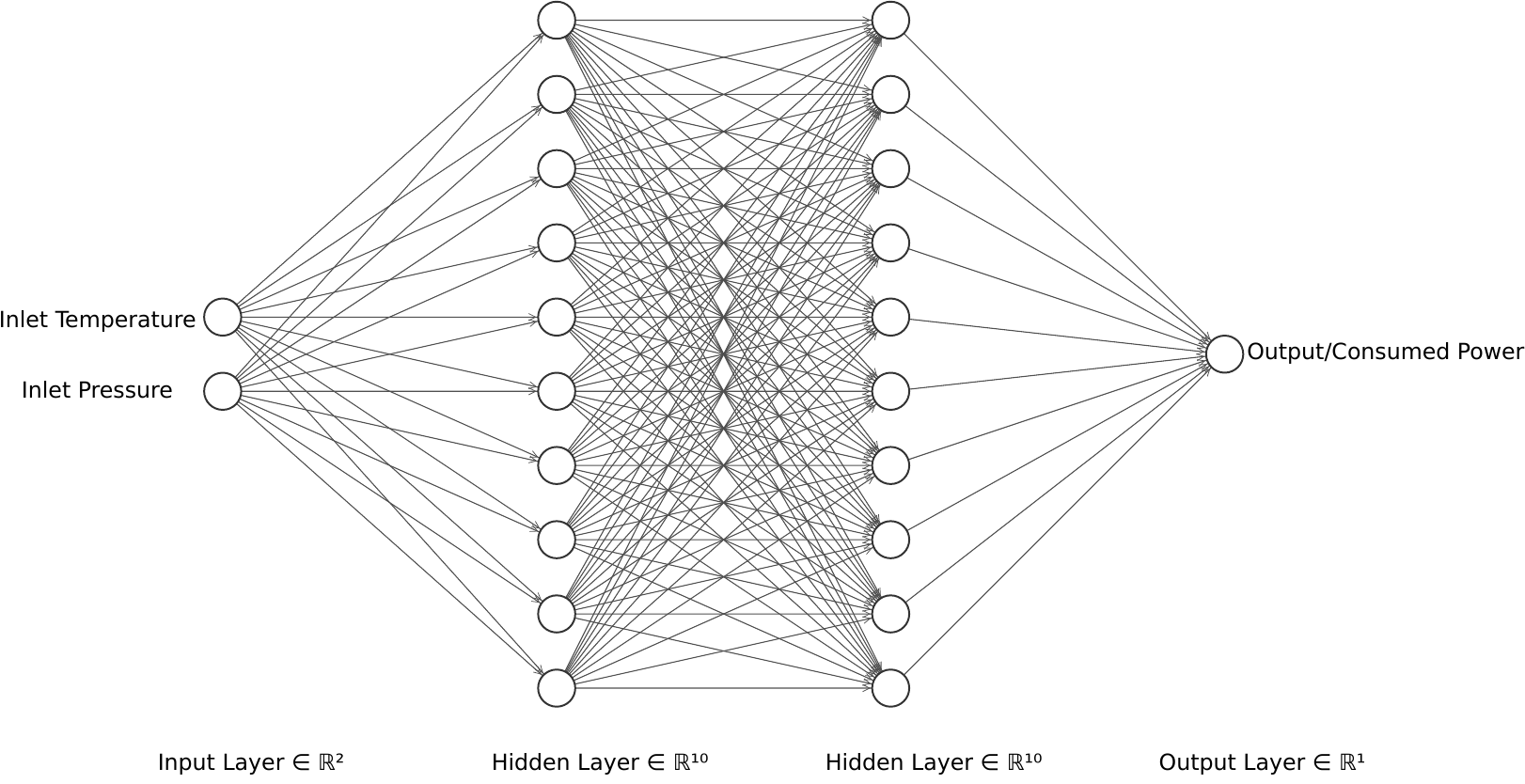}
    \caption{Feed-Forward neural network}
    \label{fig:nn}
\end{figure}

Figure \ref{fig:nn} shows the deep neural network for modeling the objectives and variables. A sequential model for the neural network were created using Tensorflow \cite{abadi2016tensorflow}. The mentioned neural network has two input neurons (inlet temperature and inlet pressure) and one output neuron (once for produced power and once for consumed power) with two hidden layers, each with ten neurons. \textit{Relu} activation function were applied for the hidden layers. After modeling the data, two-dimensional polynomial regression were used to obtain the objective functions.

\subsubsection{Genetic Algorithm}
The genetic algorithm is a heuristic optimization strategy that replicates natural evolution by changing a population of individual solutions \cite{fogel1994introduction}. Chromosomes represent design points (\(x\)). The method selects parents randomly from the existing population and uses them to produce the next generation. Since good parents produce good children, the population gradually approaches an ideal solution over successive generations. The algorithm eliminates the bad points from the generation. GA can achieve the optimal global solution without clinging to a locally optimal solution. Because GA is a probabilistic method, different runs may yield different results. As a result, many iterations are required to validate the best solution.

The genetic algorithm consists of five stages:

\begin{itemize}
    \item \textbf{Initial population:} The procedure starts with a group of data identified as a population. Each case is a potential solution to the addressed problem; Genes are a set of characteristics (variables) that describe a person. Chromosomes comprise a string of genes (solution).
    \item \textbf{Fitness function:} The fitness function defines an individual's fitness level. It assigns each case a fitness score, and the fitness score determines the likelihood of an individual being chosen for regeneration.
    \item \textbf{Selection:} This stage aims to choose the fittest individuals and pass on their genes to the next generation. Two sets of individuals are selected depending on overall fitness levels. Individuals with high fitness scores are more likely to be determined for regeneration.
    \item \textbf{Crossover:} Crossover is the most critical stage. A crossover point is a randomly selected point within the genes for each couple of parents to be matched.
    \item \textbf{Mutation:} The mutation stage is to conserve population variety and to prevent early convergence.
\end{itemize}

When the population convergence is achieved, the algorithm is terminated and the solutions are obtained.

\section{Results and Discussion}
\label{s:result}
This section will present the results and analyze the fundamental parameters of the models and compare them with each other. The grid independence test and model validation are explained in our previous work \cite{jabbary2021numerical}.

\subsection{Optimization}
\begin{figure}[!htb]
    \centering
    \includegraphics[scale=1]{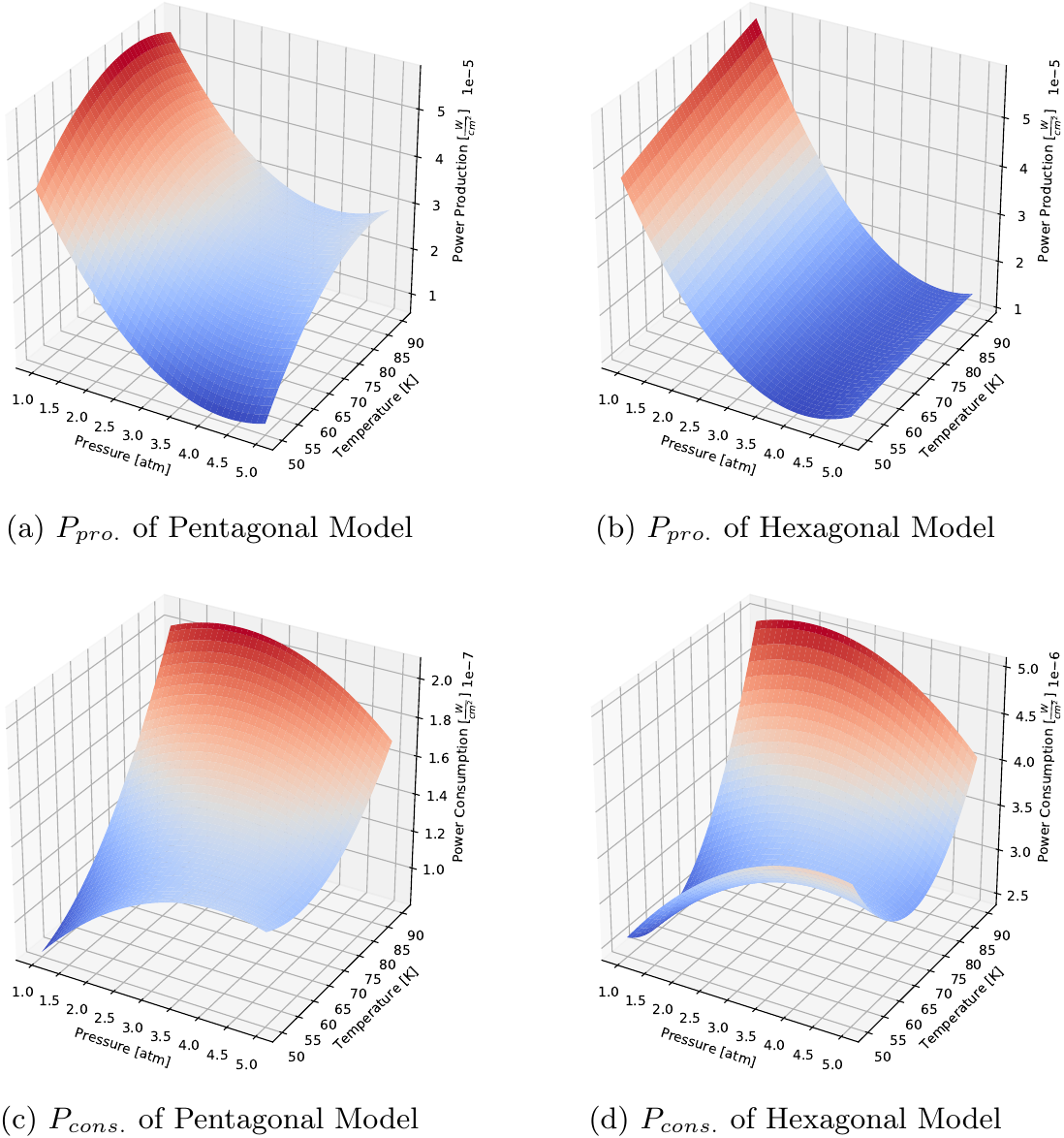}
    \caption{Changes of production/consumption power of models}
    \label{fig:surf}
\end{figure}

The desired parameters of the problem are mathematically optimized. The input parameters of this research are the inlet pressure and temperature of the cell, and the output parameters are power consumption and production power. Our goal is to maximize output power while maintaining/minimizing power consumption. To do this, the inlet pressure was changed from 1 \(atm\) to 5 \(atm\) and the inlet temperature from 50\(^\circ\)C to 90\(^\circ\)Cat the same time. Step. At each stage, consumption and production power were calculated. Figure \ref{fig:surf} Shows the changes of these powers with different pressures and temperatures in three-dimensional space.

\begin{figure}[!htb]
    \centering
    \includegraphics[scale=1]{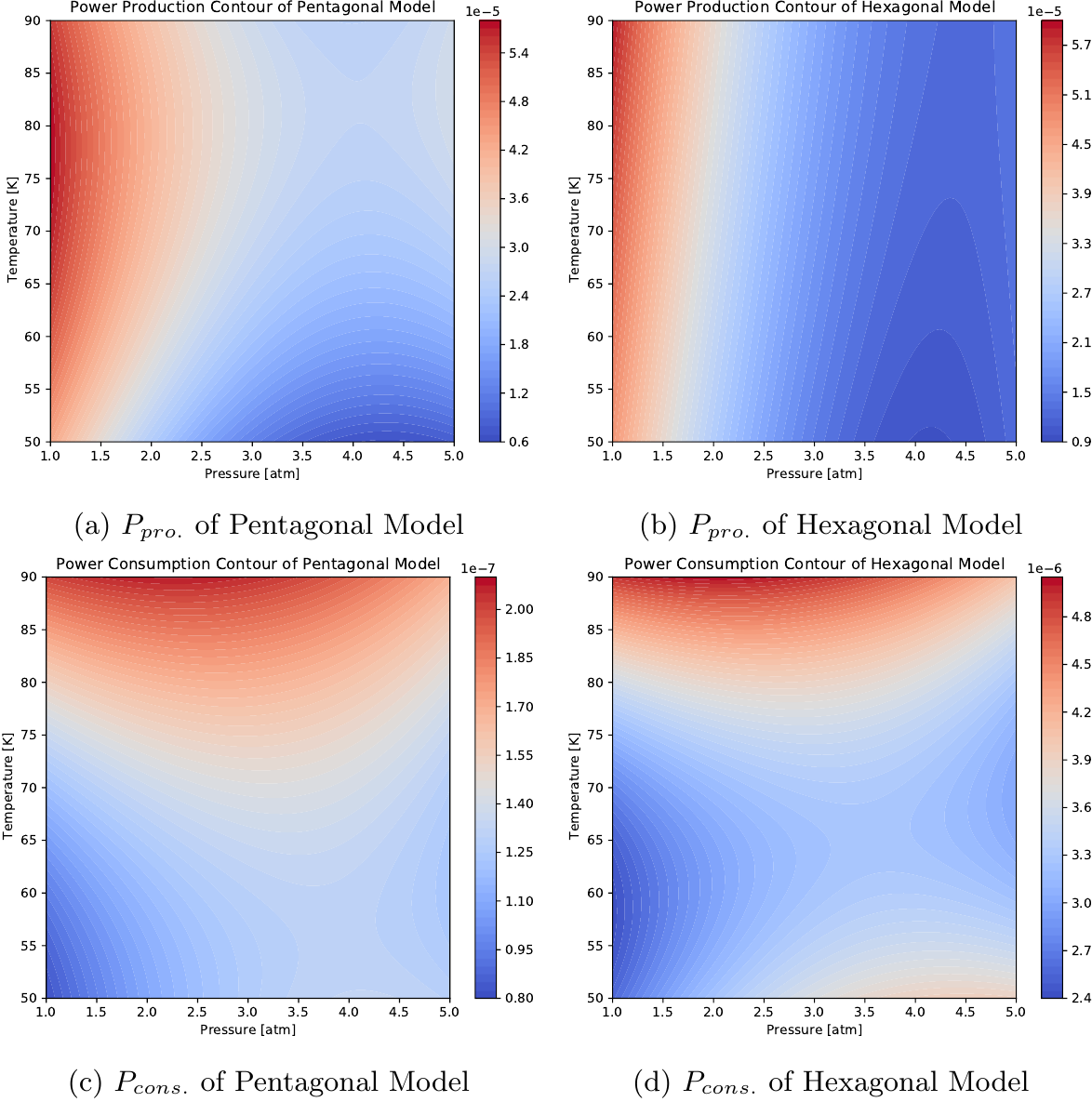}
    \caption{Contours of production/consumption power of models}
    \label{fig:contopt}
\end{figure}

Production power in both models has the highest values at all temperatures and a pressure of 1 atmosphere—the higher the inlet pressure, the lower the production power. Figure \ref{fig:contopt} shows the two-dimensional contours of these results. As shown, the power consumption in the two models, at all pressures and a temperature of 90\(^\circ\)C, has its maximum values. By decreasing the temperature, The power consumption of the fuel cell can be reduced.

To perform multi-objective optimization, The data were modeled using the neural network mentioned in Figure \ref{fig:nn}. Then, with multi-parameter polynomial regression, the mathematical relations of the problem objectives were obtained. The relationships for power production are as follows:

\textit{Pentagonal Model:}
\begin{multline}
    P_{pro} = 3.266e^{-6} P^2 + 5.816e^{-8} P T - 3.127e^{-5} P \\- 1.928e^{-8} T^2 + 2.936e^{-6} T - 3.027e^{-5}
\end{multline}

\textit{Hexagonal Model:}
\begin{multline}
    P_{pro} = 3.82e^{-6} P^2 - 6.802e^{-8}P T - 2.82e^{-5} P \\- 9.945e^{-10}T^2 + 5.052e^{-7} T + 5.251e^{-5}
\end{multline}

And the relationships for power consumption are as follows:

\textit{Pentagonal Model:}
\begin{multline}
    P_{cons} = -5.112e^{-9} P^2 - 4.847e^{-10} P T + 6.669e^{-8} P +\\ 5.415e^{-11} T^2 - 4.154e^{-9} T + 1.172e^{-7}
\end{multline}

\textit{Hexagonal Model:}
\begin{multline}
    P_{cons} = -1.111e^{-7} P^2 - 1.365e^{-8} P T + 1.68e^{-6} P +\\ 2.4647e^{-9} T^2 - 2.729e^{-7} T + 9.1835e^{-6}
\end{multline}

\begin{figure}[!htb]
    \centering
    \includegraphics{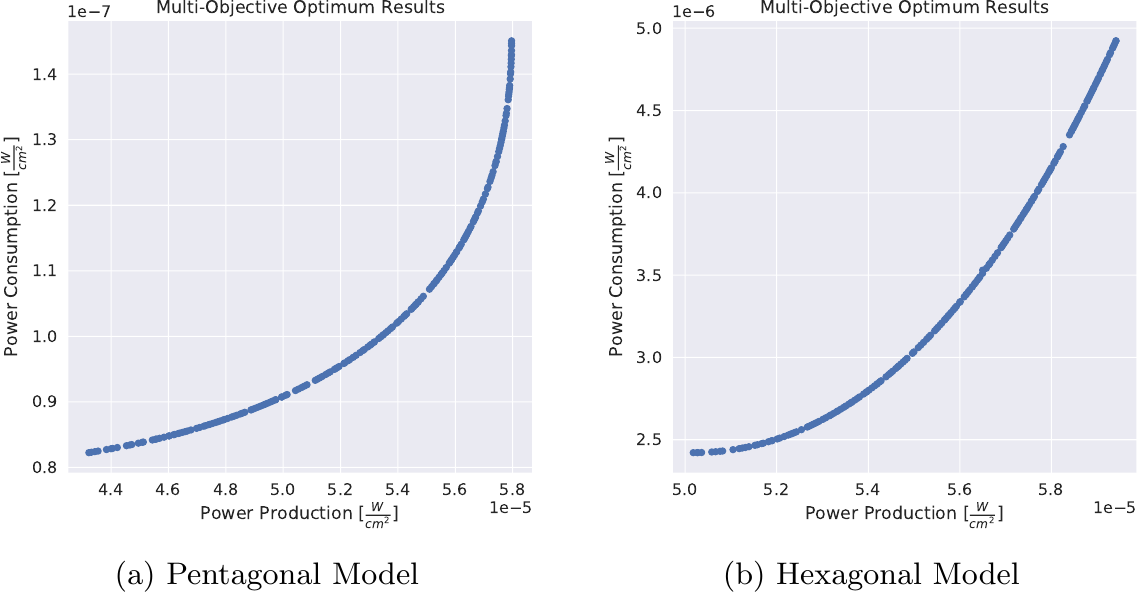}
    \caption{Comparison of optimization results}
    \label{fig:optres}
\end{figure}

After obtaining the objective functions of the problem, the \textit{NSGA-II} multi-objective genetic algorithm was used to optimize the objectives. 200 generations, population size 200 and random seed of 1 were selected. Figure \ref{fig:optres} shows the results of this optimization as the optimal range between the two objective functions. According to this figure, in both models, the higher the production power, the higher the fuel cell's consumption power. To find reasonable values for the input parameters of the problem, these results shoulld be examined. In both models, the device's power consumption is much less than the production power. This value is 0.198\% of the production power on average in the Pentagonal model and 6.21\% of the production power on average in the Hexagonal model. Considering this, if the maximum production power is available in both models, the power consumption in the Pentagonal model is only 0.25\% of the production power.
In comparison, the power consumption in the Hexagonal model is 8.29\% of the production power. As a result, if the input parameters are used to achieve the maximum output power, the power consumption will still be much lower. In this case, the inlet pressure and temperature in the Pentagonal model will be 1 atm and 77.645 \(^\circ\)C, and in the Hexagonal model, 1 atm and 90 \(^\circ\)C, respectively.

\subsection{Polarization Curves and Current Density}
\begin{figure}[!htb]
	\centering
	\subfloat[Part 1]{\includegraphics[scale=0.32]{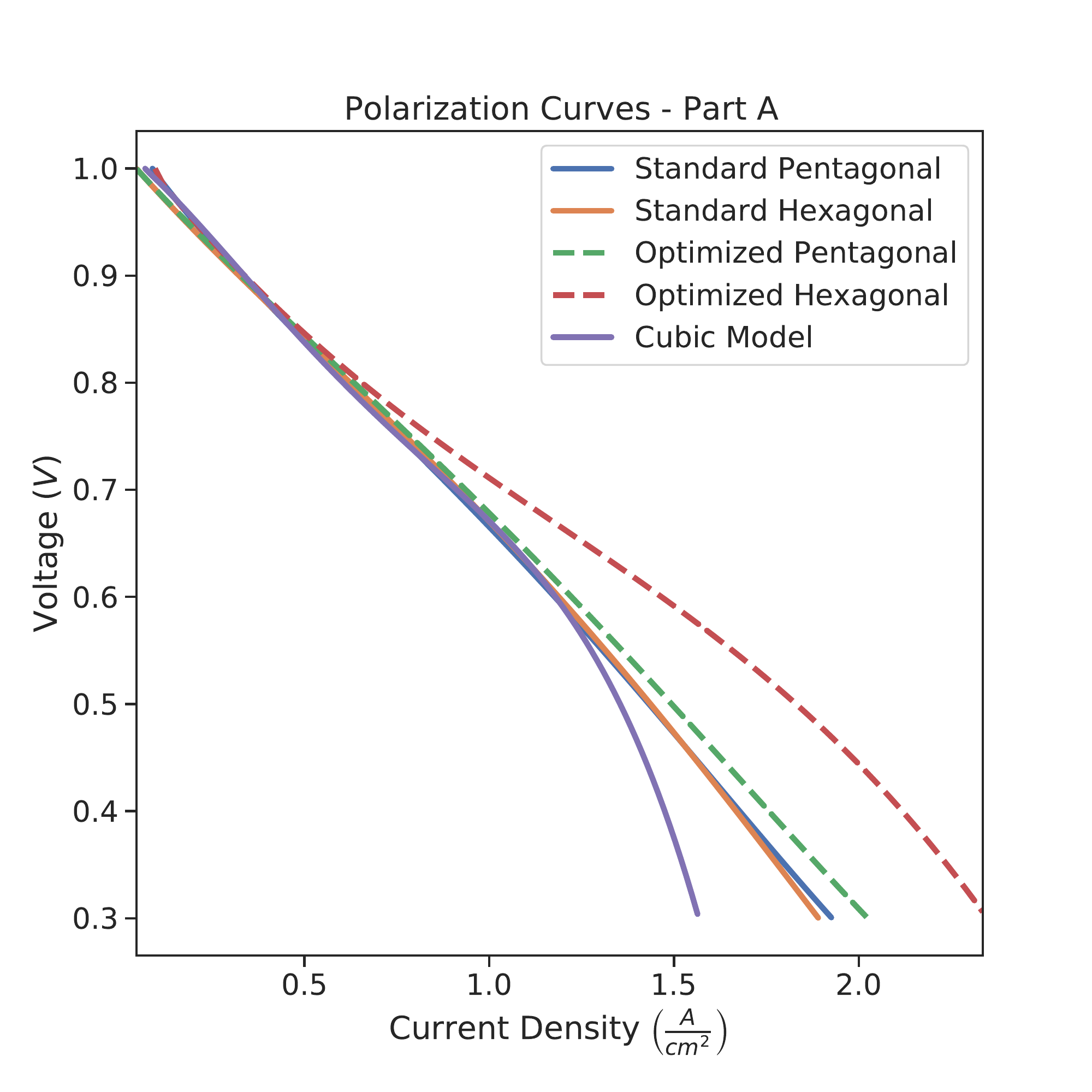}}
	\subfloat[Part 2]{\includegraphics[scale=0.32]{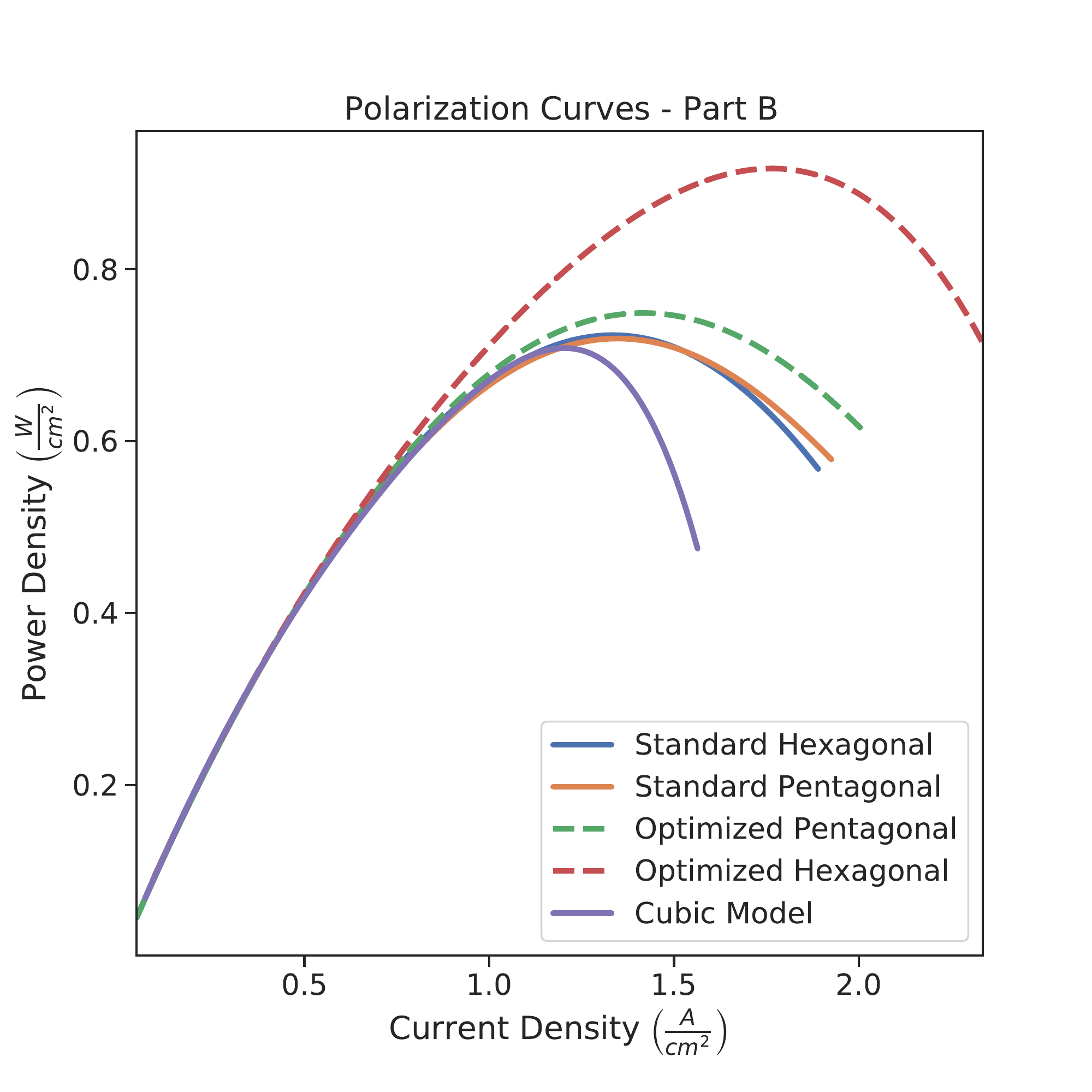}}
	\caption{Polarization Curves of the Models}
	\label{fig:pol1}
\end{figure}

\begin{figure}
    \centering
    \includegraphics[scale = 0.6]{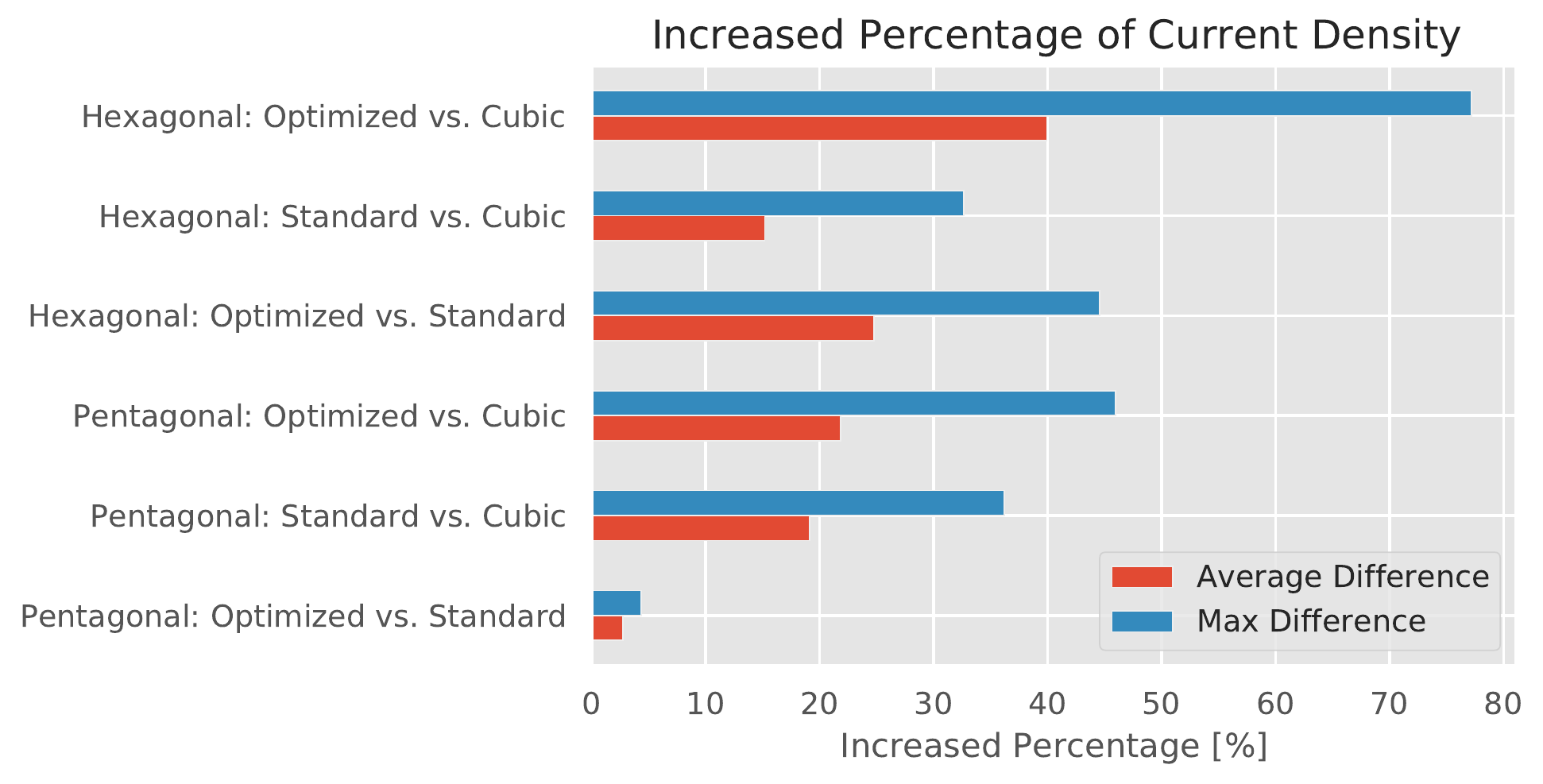}
    \caption{Increased percentage of current density of the presented models}
    \label{fig:diffbar}
\end{figure}

Fig \ref{fig:pol1} shows the polarization curves of the presented models. The presented models produced more current and power density than the Cubic model. The performance of the optimized models in this field is more than the standard models due to optimized parameters. To be able to compare these values accurately, Table \ref{tab:diff} and Figure \ref{fig:diffbar} indicate the percentage increase in the current density of the proposed models.

First, standard models perform better than base models. The Pentagonal and Hexagonal models have an average current density of 19.096\% and 15.179\% higher than the base model. Second, the current densities generated by the optimized models are even higher than the standard models, which are shown in Figure \ref{fig:diffbar}. The optimized pentagonal and hexagonal models have an average current density of 21.819\% and 39.931\% higher than the base model.

Two main reasons for these enhancements are optimal parameters and the new cell design. The performance of the presented models at near-open voltage is relatively similar to the cubic model. However, at low voltages, their performance has significantly enhanced.

\begin{table}
	\centering
	\caption{Differences of output current densities between models}
	\label{tab:diff}
	\begin{tabular}{|l|l|l|l|} 
		\hline
		~                                  & Max diff. & Avg. diff. & Unit \\ 
		\hline
		Pentagonal: Optimized vs. Standard & 4.3            & 2.7             & \%   \\ 
		\hline
		Pentagonal: Standard vs. Cubic     & 36.2           & 19.1            & \%   \\ 
		\hline
		Pentagonal: Optimized vs. Cubic    & 46             & 22              & \%   \\ 
		\hline
		Hexagonal: Optimized vs. Standard  & 44.5           & 24.7            & \%   \\ 
		\hline
		Hexagonal: Standard vs. Cubic      & 32.6           & 15.2            & \%   \\ 
		\hline
		Hexagonal: Optimized vs. Cubic     & 77.2           & 40              & \%   \\
		\hline
	\end{tabular}
\end{table}

\subsection{Effects of Relative Humidity}
\begin{figure}[!htb]
	\centering
	\subfloat[Pentagonal Model]{\includegraphics[scale=0.32]{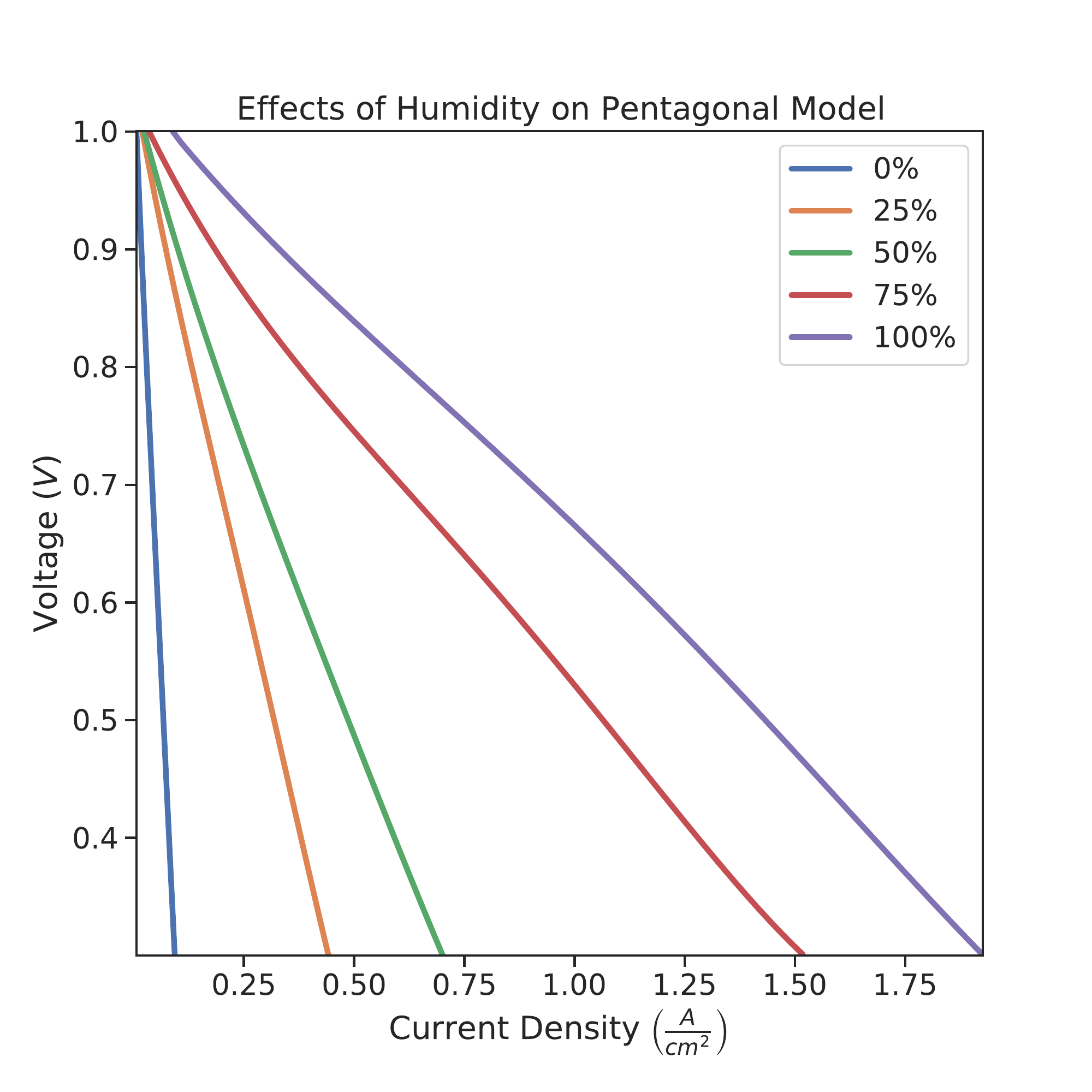}}
	\subfloat[Hexagonal model]{\includegraphics[scale=0.32]{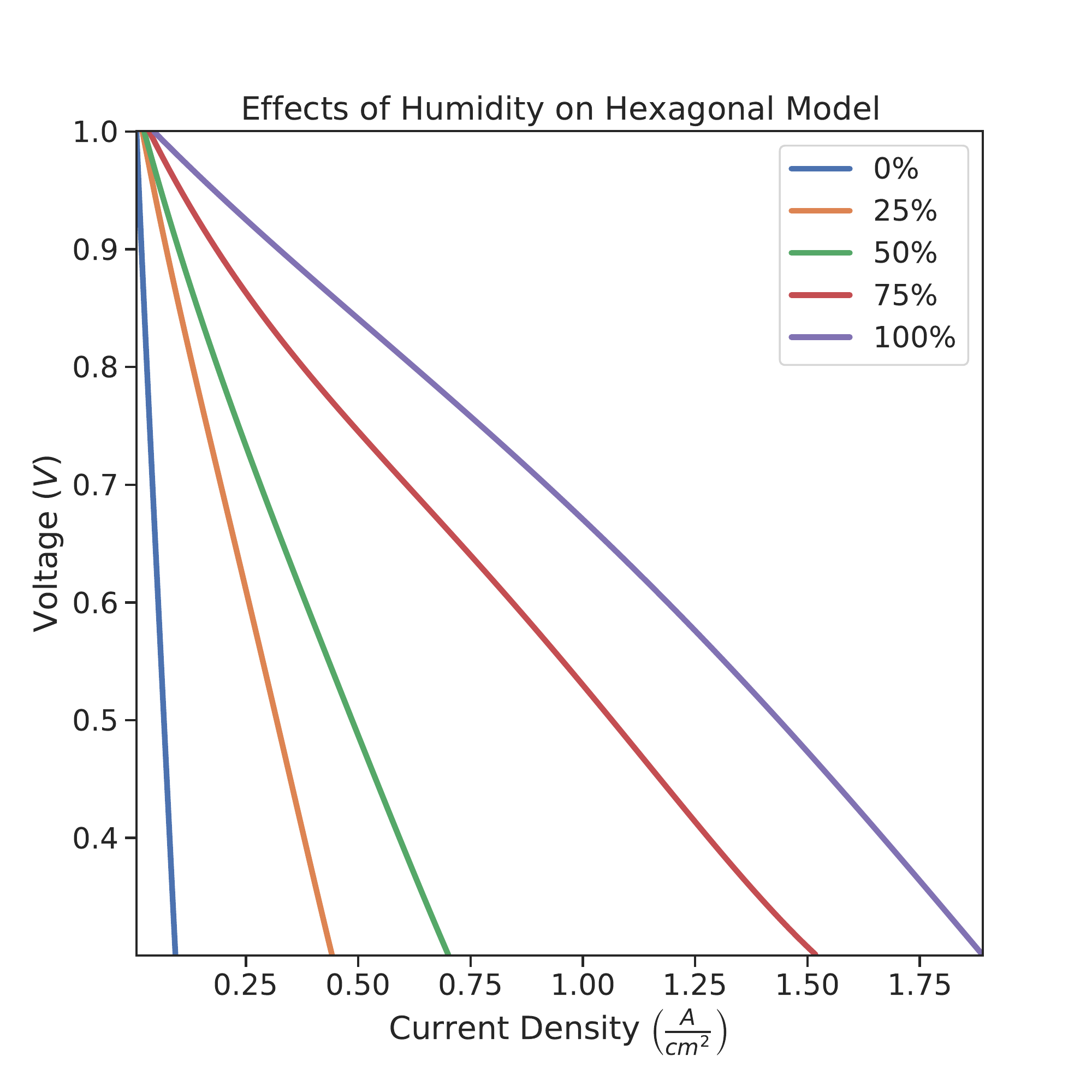}}
	\caption{Effects of humidity changes on models}
	\label{fig:hum}
\end{figure}

Figure \ref{fig:hum} displays the effect of Relative Humidity (RH) on the performance of the standard presented fuel cell models. Humidity is among the main factors influencing fuel cell performance. Decreasing RH causes a reduction in the membrane's proton transfer conductivity in both cases. Overall, decreasing RH may decrease electrode kinetics, including electrode reaction and mass diffusion rates and membrane proton conductivity, leading to a severe decrease in cell efficiency. According to this result, 100\% humidity is reported to achieve adequate performance for both models.

\subsection{Liquid Water Content}
\begin{figure}[!htb]
	\centering
	\subfloat[Pentagonal Model]{\includegraphics[scale=0.32]{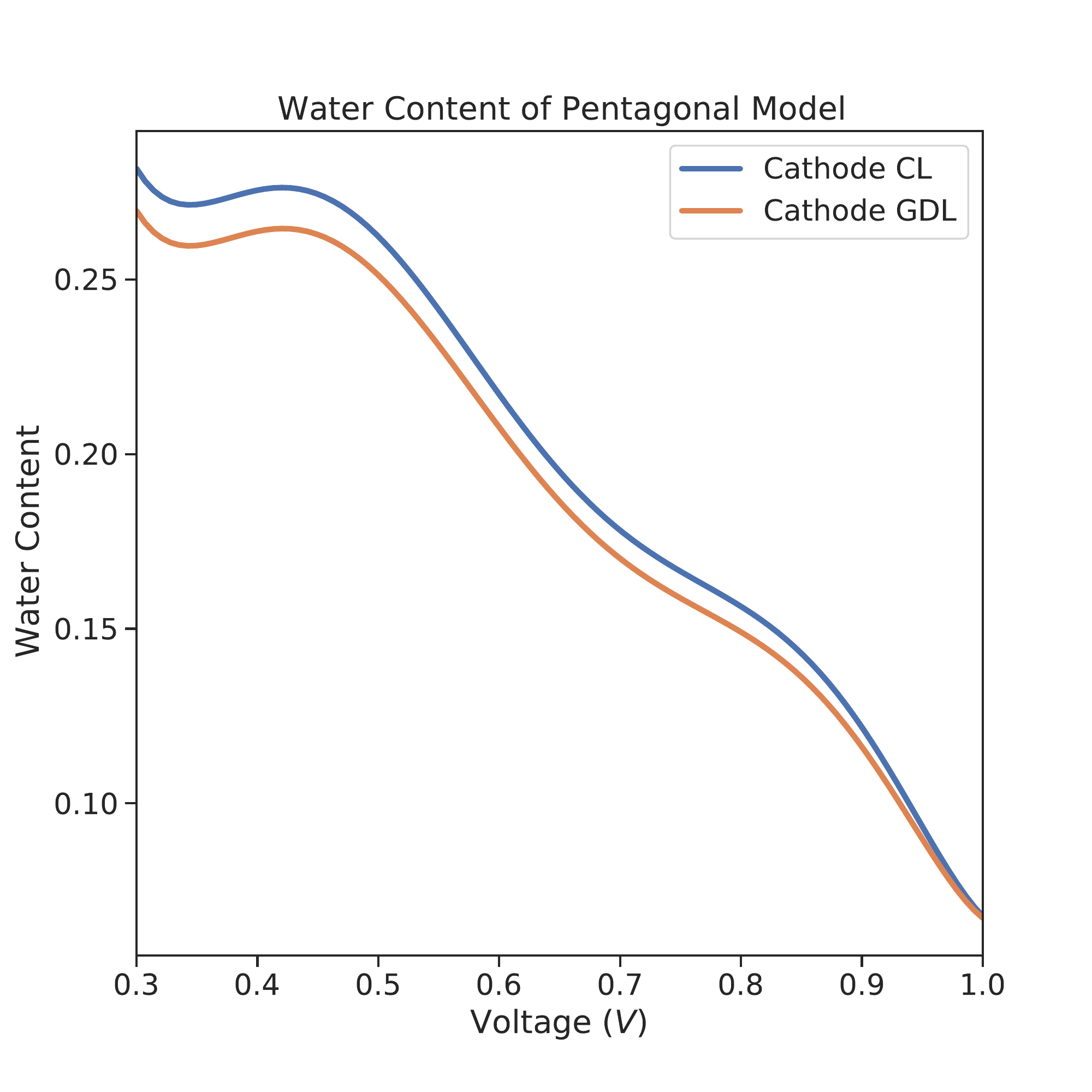}}
	\subfloat[Hexagonal model]{\includegraphics[scale=0.32]{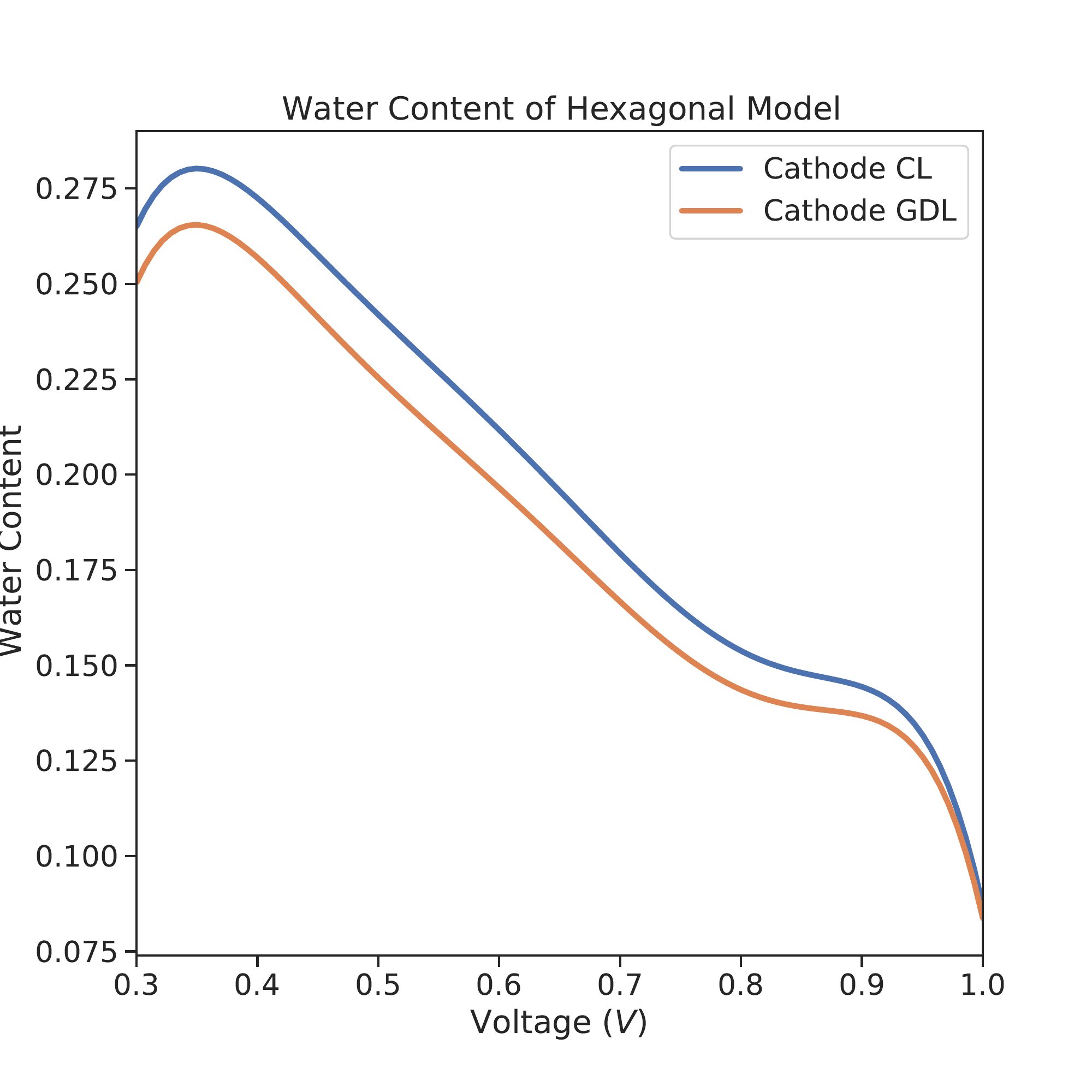}}
	\caption{Effects of humidity changes on models}
	\label{fig:wat}
\end{figure}

The amount of liquid water in the MEA influences the proton conductivity and the activation overpotential. If MEA is not sufficiently hydrated, proton conduction drops, and cell resistance increases. On the other hand, surplus water can cause problems in fuel cells, including water flooding.
Figure \ref{fig:wat} shows the volumetric average of water content in the layers of the standard presented models. Since This content is minimal in the anode part, the cathode sections' water content were displayed.
The water content at humidities below 100\% is minimal in the fuel cell layers. This reduction in water content may cause the membrane to dry out and ultimately reduce the efficiency of the fuel cell.

The amounts of water in both models are close to each other. At 0.355V and 0.375V, the cathode layers of both the hexagonal and pentagonal models contain the highest water content.

\section{Conclusion}
\label{s:conc}
This paper presented two new designs for PEM fuel cells with pentagonal and hexagonal shapes. After obtaining an increase in the performance of these models compared to the cubic model in generating current density and electrical power, they were optimized. The data were trained using neural networks and regression techniques. The Response Surface Method \textit{(RSM)} was applied to derive the mathematical function corresponding to the problem objectives: production and consumption power. Using a multi-objective genetic optimization algorithm (\textit{NSGA-II}), the targets were simultaneously optimized for problem inputs including operating temperature and pressure. Having the optimal values for these models, the optimized results were compared with the results of standard models. In addition, the effects of changes in the relative humidity of the cell inlet in the models were analyzed and the liquid water content was investigated.

The optimal designs outperform the base model (Cubic Model) and are more effective than the standard models. The average increase in output current density of the optimal models compared to the base model (Cubic) is 21.819\% and 39.931\% in the pentagonal and hexagonal models. Compared to optimized and standard cases, the mentioned percentage is 2.722\% and 24.752\% in the pentagonal and hexagonal models. The effect of relative humidity (RH) on the channel inputs of standard models were investigated. 100\% humidity is the optimal setting for achieving the highest possible current density in the models. Reducing the relative humidity of the inlet causes the MEA to dehydrate, reducing the current density and ultimately reducing the cell efficiency. The average volumetric volume of liquid water content was measured in the cathode section layers. For standard pentagonal and hexagonal models, the corresponding voltages for the highest liquid water contents are 0.379V and 0.355V.

\bibliographystyle{elsarticle-num-names}
\bibliography{ref}

\end{document}